\documentclass[journal]{IEEEtran1}
\usepackage{times}  
\usepackage{helvet}  
\usepackage{courier}  
\usepackage{url}  
\usepackage{graphicx}  
\usepackage{amssymb}
\usepackage[numbers]{natbib}
\usepackage{xcolor}
\usepackage{amsmath}
\usepackage{textcomp}
\usepackage{siunitx}
\usepackage{graphicx}
\usepackage{caption}
\usepackage{booktabs}
\usepackage{multirow}
\usepackage{tabularx}
\usepackage{subcaption}
\usepackage{tikz} 
\usepackage[ruled,vlined]{algorithm2e}
\usepackage{graphicx,color}
\usepackage{multicol}

\usepackage{multirow}
\usepackage{subcaption}
\usepackage{algorithmic}
\usepackage{threeparttable}
\usepackage{booktabs,caption}
\usepackage{threeparttable, tablefootnote}
\usepackage[ruled,vlined]{algorithm2e}
\usepackage{etoolbox}

\usepackage{soul}
\usepackage{xcolor}

\ifCLASSINFOpdf
 
\else
 
\fi

\begin{document}

\title{Deep Inverse Feature Learning: A Representation Learning of Error}

\author{Behzad Ghazanfari, Fatemeh Afghah\\
School of Informatics, Computing, and Cyber Security, \\Northern Arizona University, Flagstaff, AZ 86001, USA. 
}

\maketitle
\begin{abstract}

This paper introduces a novel perspective about error in machine learning and proposes \textit{inverse feature learning (IFL)} as a  representation learning approach that learns a set of high-level features based on the representation of error for classification or clustering purposes. The proposed perspective about error representation is fundamentally different from current learning methods, where in classification approaches they interpret the error as a function of the differences between the true labels and the predicted ones or in clustering approaches, in which the clustering objective functions such as compactness are used. Inverse feature learning method operates based on a deep clustering approach to obtain a qualitative form of the representation of error as features. The performance of the proposed IFL method is evaluated by applying the learned features along with the original features, or just using the learned features in different classification and clustering techniques for several data sets. The experimental results show that the proposed method leads to promising results in classification and especially in clustering. In classification, the proposed features along with the primary features improve the results of most of the classification methods on several popular data sets. In clustering, the performance of different clustering methods is considerably improved on different data sets. There are interesting results that show some few features of the representation of error capture highly informative aspects of primary features.  We hope this paper helps to utilize the error representation learning in different feature learning domains. 

\end{abstract}


\IEEEpeerreviewmaketitle

\section{Introduction}

Error as the main source of knowledge in learning has a unique and key role in machine learning approaches. In current machine learning techniques, the error is interpreted without considering its representation in a variety of forms in objective functions, loss functions, or cost functions. In supervised approaches, different loss functions and cost functions have been defined to measure the error as the differences between the true and the predicted labels during the training process in order to provide decision functions which minimize the risk of wrong decisions \citep{rosasco2004loss}. In clustering, high-level objectives in the form of clustering loss (e.g., compactness or connectivity) are used to consider similar instances as the clusters. The clustering objectives minimize the cost functions that are defined as the distances between the instances and the representative of clusters (e.g., the centers of clusters in k-means \citep{lloyd1982least}). In this paper, we define the error in a more informative format which captures different representations of each instance based on its class or cluster and uses that toward learning several high-level features to represent the raw data in a compact way. 

The proposed \textit{error representation learning} offers a new approach in which the error is represented as a dynamic quantity that captures the relationships between the instances and the predicted labels in classification or the predicted clusters in clustering. This method is called ``\textit{inverse feature learning (IFL)}'' as it firsts generates the error and then learn high-level features based on the representation of error. We propose IFL as a framework to transform the error representation to a compact set of impactful high-level features in different machine learning techniques. The IFL method can be particularly effective in the data sets, where the instances of each class have diverse feature representations or the ones with imbalanced classes. It can also be effective in active learning applications with a considerable cost of training the instances.

In summary, the error representation learning method offers a different source of knowledge that can be implemented by different learning techniques and be used alongside a variety of feature learning methods in order to enhance their performance. In this paper, we present a basic implementation of this strategy to learn the representation of error to focus on the concept, its potential role in improving the performance of state-of-the-art machine learning techniques when a general set of high-level features are utilized across all these techniques. Indeed, the proposed technique can be improved by using additional feature representations as well as defining other dynamic notions of error. The experimental results confirm that even a few simple features defined based on error representation can improve the performance of learning in classification and clustering.

\section{Related Works} \label{sec:Related_Works}
The error in most machine learning approaches, depending on whether the approach is supervised or unsupervised has a similar definition of the difference between the predicted and true labels in supervised ones or the clustering losses in unsupervised ones. Also, the term error refers to the same concept in similarity learning \citep{koestinger2012large}, semi-supervised learning \cite{chapelle2009semi}, self-supervised learning \citep{de1994learning}.
Generative methods are distinguished from discriminative methods in machine learning in terms of considering the underlying distribution, but the error is the same depend on is that the approach is supervised or unsupervised.

Representation learning methods are known for their role in improving the performance of clustering and classification approaches by learning new data representations from the raw data which better presents different variations behind the data \citep{lecun2015deep}. Classification with representation learning typically is a closed-loop learning with many different architectures. Clustering with deep learning, while does not use the labels, can lead to promising results \citep{min2018survey}. Unsupervised representation learning methods can be used for pre-training nets, generally for deep networks, or for extracting high-level features, denoising, and dimensionality reductions.

Representation learning approaches in supervised or unsupervised applications are generally based on elements such as restricted Boltzmann machines (RBMs) \citep{smolensky1986info}, autoencoder \citep{bourlard1988auto,hinton1994autoencoders}, convolutional neural networks (ConvNets) \citep{lecun1998gradient}, and clustering methods \citep{min2018survey}. In the majority of existing classification with representation learning methods, the term error refers to a function of the differences between the true and the predicted labels (i.e., loss functions) \citep{janocha2017loss}. Also, the error in clustering with representation learning methods is defined as clustering loss of high-level objectives alongside other loss functions \citep{aljalbout2018clustering}. 

In unsupervised feature learning such as autoencoders, the error still is defined in the form of reconstruction error between the input data and the resultant of encoding and decoding process \citep{bourlard1988auto,hinton1994autoencoders}. Regularization terms \citep{kukavcka2017regularization} or dropout \citep{srivastava2014dropout} are used alongside with loss functions to enhance the training process to have a better generalization of the learned features (e.g., learning the weights of neural nets). Generative adversarial nets (GANs) \citep{goodfellow2014generative} use two separate neural networks competing against each other with the ultimate goal of generating synthetic data similar to the original inputs through the generator. However, the concept error in GANs is the same as the other ML approaches. In this paper, we propose a novel notion of error as the resultant representation of assigning the data instances to different available classes or clusters as a source of knowledge for learning. We would like to note that this approach is different from generative approaches such as GANs which are based on data generation while the proposed method is based on the virtual assignment of the data instances to different classes or clusters to consider the resultant representation of such assumptions for each instance in relation to the existing classes or clusters. 

This proposed method is also different from similarity learning \cite{koestinger2012large}, which learns a similarity function to measure how similar two objects are, in the sense that it extracts the relationships between the objects and the classes. Self-supervised learning methods are based on finding patterns inside of input instances \cite{de1994learning}. The semi-supervised learning and active learning methods, that utilize a combination of classification and clustering, are based on the assumption that the instances which are in the same cluster have the same label and using this assumption toward predicting the labels for new instances  \cite{seeger2000learning,chapelle2003cluster, chapelle2009semi,benabdeslem2014efficient}. In these methods, the instances near the center of clusters are considered as the most representative objects for the purpose of determining the labels. Other approaches including \cite{xu2003representative,nguyen2004active} utilized clustering for active learning in several different ways.

In summary, the proposed method is different from existing techniques in the literature since they focus on data representation learning and calculate errors in classification by the loss functions or cost functions to minimize the risk by taking the difference between the predicted labels and true labels. In clustering with deep learning, the objectives are defined in the form of clustering loss and auxiliary losses \citep{min2018survey} which are still based on data representation. The distinctions of the inverse feature learning related to common trends in machine learning is that the inverse feature learning method generates the error by a trial and calculates the resultant representation as the error and then transform the error as high-level features to represent the characteristics of each instance related to the available classes or clusters.

\section{Notations}\label{Notations}

First, we define the notations which are used for both clustering and classification. Let us consider $X$ and $Y$ to refer to the input and the output spaces, respectively in which each instance $x_i \in X$ consists of $h$ features --- $x_{i}=\langle x_{i,1}  , \cdots,x_{i,h} \rangle$. We use notation $s$ to denote the number of classes or the number of clusters.  We consider $C$ to denote the set of clusters in clustering, $C=\langle c_{1}, \cdots, c_s\rangle$, as $\forall x_{j} \in X,\ \exists c_{i} \in C\ |\ x_{j} \in c_{i}$. $Y$ is the corresponding classes of instances in the classification. 
The set of clusters, $C$ can simply correspond to $Y$ in  assignment problems using approaches such as Hungarian algorithm.

Notation $\mu_i$, $i \in \{1,...,s\}$ refers to the center of cluster $i$. $|b|$ indicates the number of instances in set $b$. In continue, we define the specific notations used for the classification and clustering approaches.

In clustering, the input data set is presented with $D=\langle X \rangle$, in which  $X=\{x_{1},\cdots, x_{n}\}$  indicates the set of input instances and $n$ shows the number of input instances --- $|X|=n$. 

In classification, the input training data set is presented with $D=\langle X^{Train}, Y^{Train} \rangle$, in which  $X^{Train}=\{x_{1},\cdots, x_{n_{train}}\}$  indicates the set of input training instances and $n_{train}$ shows the number of input instances in the training partition. The label set is denoted by $Y^{Train}$, where $Y^{Train}=\{y_{1},\cdots, y_{n_{train}}\}$ is a vector corresponded with data set $X^{Train}$. Thus, $y_{i}$ shows the corresponding label for $x_{i}$.  $X^{\text{Test}}=\langle x^{'}_{1},\cdots,x^{'}_{n_{test}} \rangle$ denotes the test set in which $n_{test}$ refers to the number of test instances. --- $|X^{Train}|=n_{train}$ and $|X^{Test}|=n_{test}$.

$Z$ refers to the latent feature space, in which $z_i=<z_{i,1},\cdots,z_{i,e}>$ is the new representation of $x_i$, and $e$ denotes the dimension of latent feature space, where the dimension of the latent space is usually much smaller than the original space (i.e.,  $e<<h$). 

\section{Deep Inverse Feature Learning}

\begin{figure}
\begin{subfigure}{1\columnwidth}
\centering
\includegraphics[width=1\columnwidth,height=3cm]{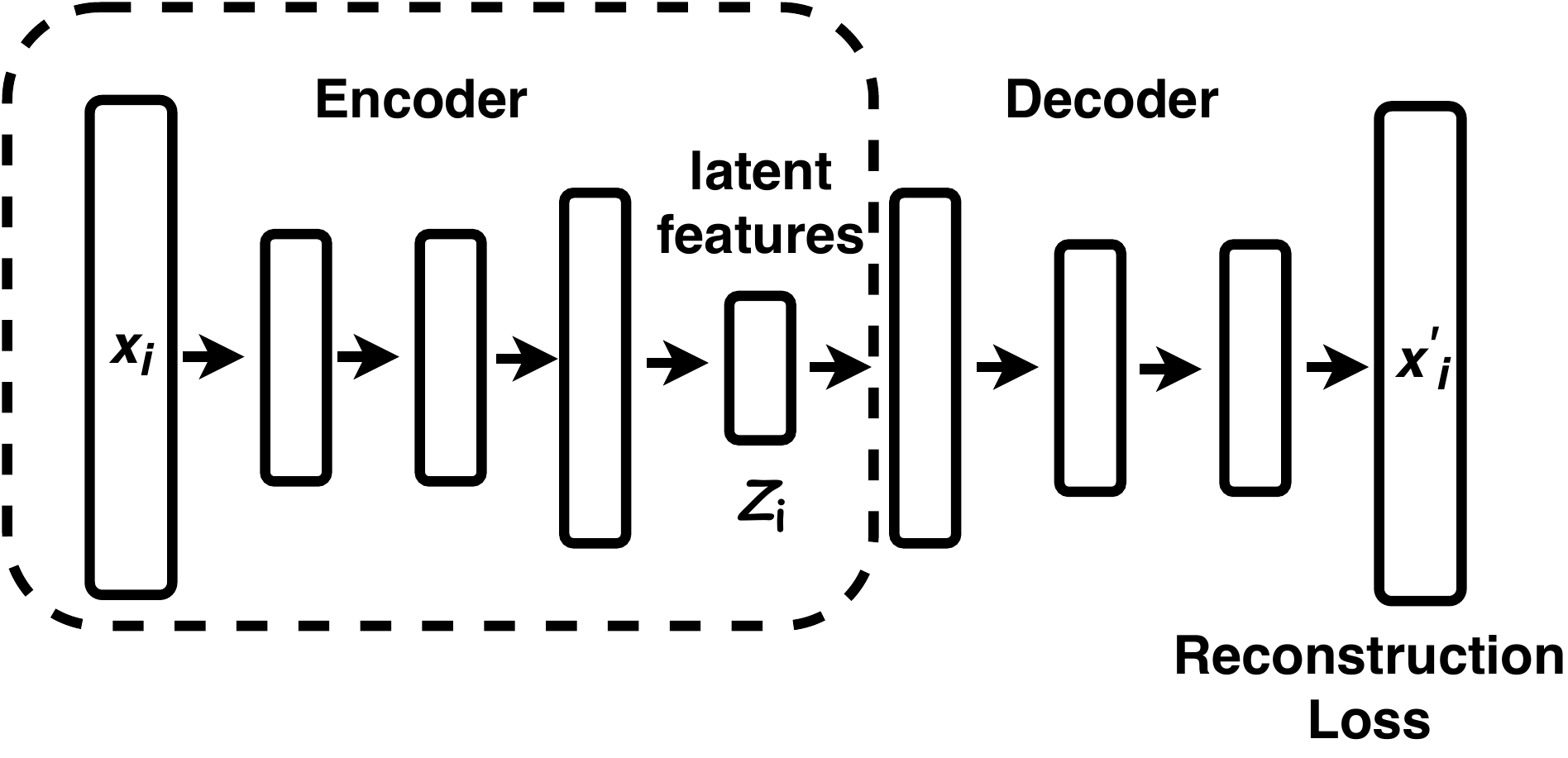}
    \caption{A simple autoencoder which learns a latent feature space, $Z$, and provides initial clusters' centroids as the first phase of DEC. The hashed part is used in the second phase of DEC after the training phase.}
    \label{fig:diagram_DEC_ae}
\end{subfigure}

\begin{subfigure}{1\columnwidth}
\centering
\includegraphics[width=1\columnwidth,height=4cm]{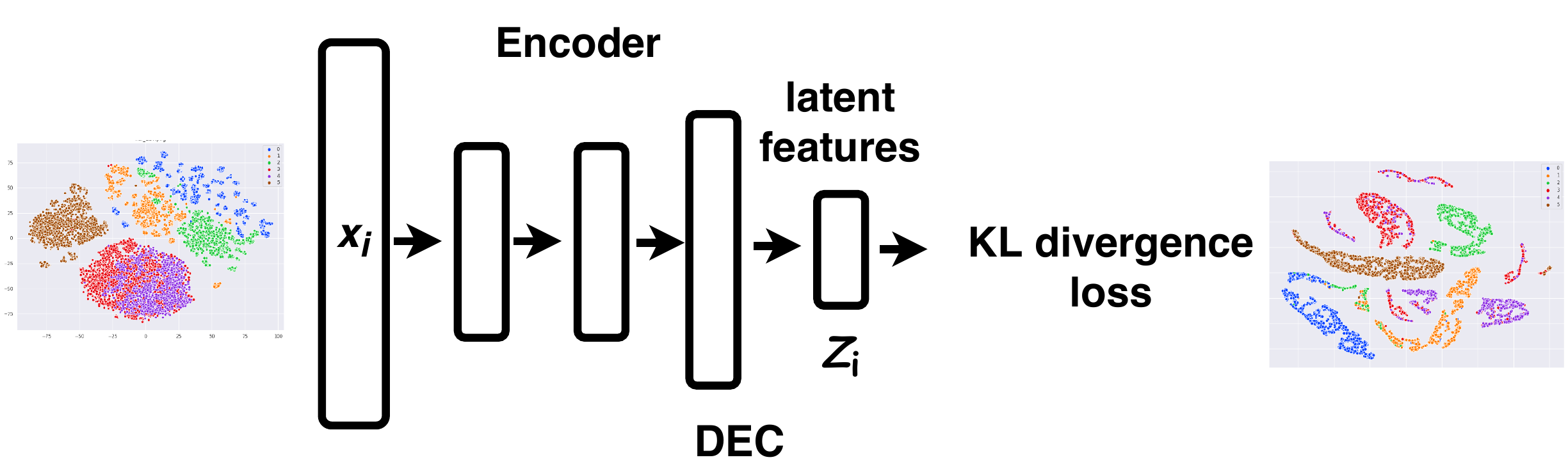}
    \caption{In the second phase of DEC architecture, a clustering based on KL divergence metric is repeated untill the method converges.}
    \label{fig:diagram_second_part_DEC}
\end{subfigure}
 \caption{ The network structure of deep embedded clustering (DEC)\citep{xie2016unsupervised}.}
\label{fig:DEC}
\vspace{-0.4 cm}
\end{figure}

In this section, we describe the proposed inverse feature learning, which is defined as a trail process that learns a set of features of the error representation from the latent features of the instances using a deep clustering approach. The learned features of the representation of error can be used for clustering or classification purposes. Here, we describe how the representation of error is generated, and then presented in the form of learned features. First, we provide a high-level review of different steps of this method and then describe them in more detail in the following sub-sections. The pseudo-codes for the IFL in clustering and classification are presented in Algorithm \ref{alg:IFL_cluster}, and Algorithms \ref{alg:IFL_class}, \ref{alg:IFL_class_train}, and \ref{alg:IFL_class_test}, respectively.

Clustering with deep learning is a practical and efficient solution to handle different data sets, especially the high dimensional ones. Autoencoders can provide a latent feature space which is much smaller than the original feature set \citep{aljalbout2018clustering}. Thus, a clustering method based on an autoencoder has a proper foundation for the proposed inverse feature learning. There are several types of deep clustering methods which can be categorized based on different factors such as the architecture, being deterministic or generative, or the loss functions \citep{aljalbout2018clustering,min2018survey}. We use deep embedded clustering (DEC) as a general and basic approach in the literature of clustering with deep learning that simultaneously learns feature representations and cluster assignments \citep{xie2016unsupervised}. The autoencoder in DEC offers a robust and embedded representation of data with a nonlinear mapping of the original features to a latent feature space $Z$. DEC is deterministic, and uses an autoencoder  network with learnable parameters based on the fully connected networks (FCNs) and the reconstruction loss in its first phase. Then, the DEC clusters the latent space to $s$ clusters by using cluster assignment hardening in its second phase to generate the centroid, $\mu$ of the clusters. The network structure of DEC is shown in Figure \ref{fig:DEC}. 

\begin{figure}
\centering
\includegraphics[width=0.9\columnwidth,height=2.9cm]{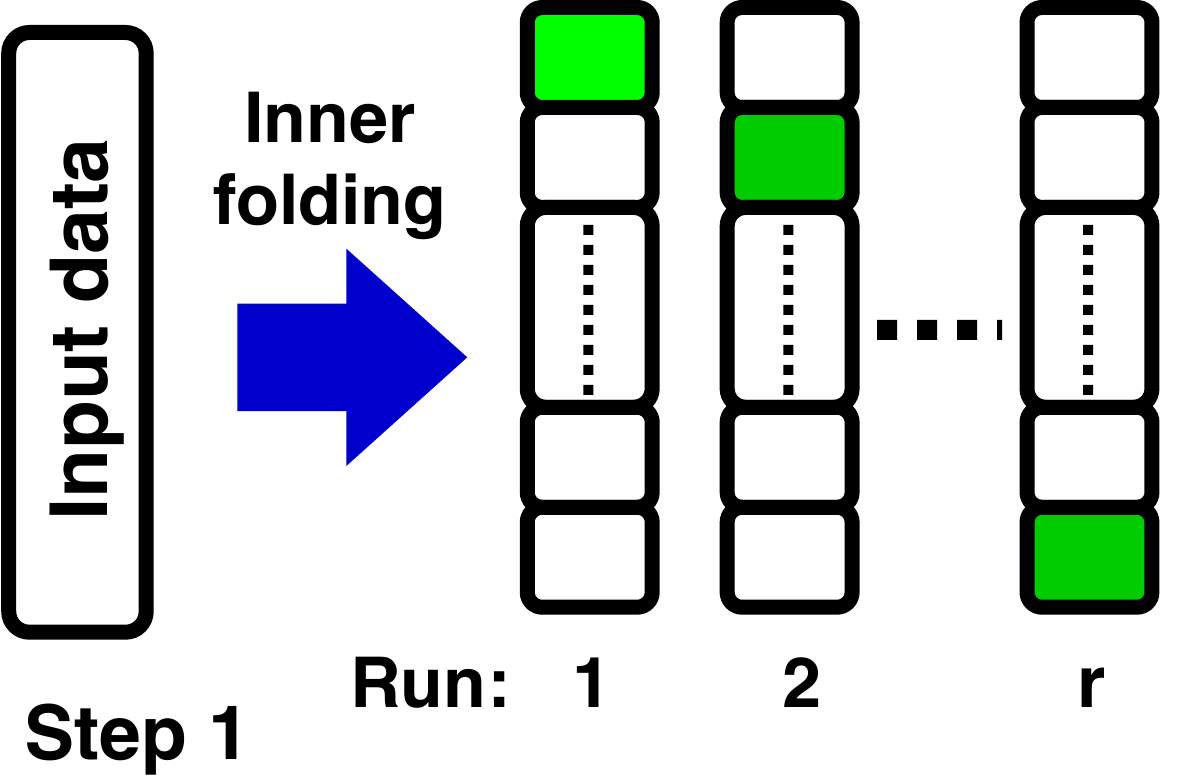}
    \caption{The inner folding process. Step 1: partition the input data to $r$ non-overlapping folds or partitions. A loop with $r$ runs is applied, where in each run, one fold is considered as an inner test and $r-1$ remaining folds are considered as inner-train. The inner-test fold in each run is colored with green.}
    \label{fig:Inner_folding_process}
\end{figure}

\begin{figure}
\centering
\includegraphics[width=1\columnwidth,height=7cm]{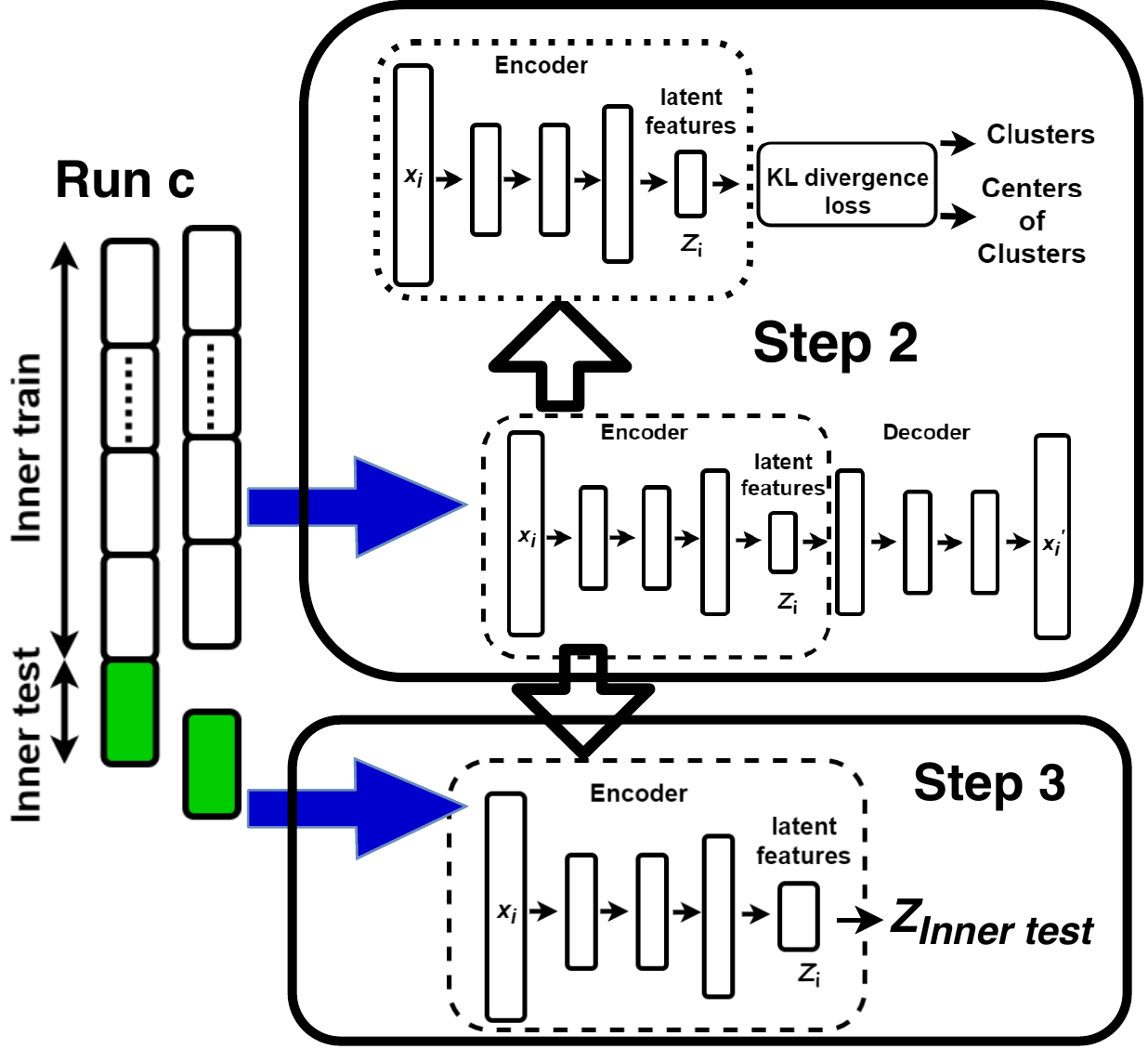}
    \caption{Steps 2 and 3: \textit{Learning the clustered representation of the inner train data} as step 2 by using the DEC to train the autoencoder in the first phase and cluster the latent feature space and obtain their centers in the second phase. The trained encoder in the first phase is used in the second phase of the DEC and also in the \textit{extracting the latent features of inner test} as step 3.}
    \label{fig:diagram_23}
    \vspace{-0.1 cm}
\end{figure}

The inverse feature learning is based on four main steps: 

\noindent \textbf{Step 1- Inner folding }: We introduce inner folding as a method to perform a trial process to extract the desired characteristics of error for all the instances. This process is inspired by $k$-fold cross-validation, where the data set is partitioned to $r$ folds. One partition of data is considered as \textit{inner test} and other partitions as \textit{inner train} in each run. However, the objective of the inner folding process is different from cross-validation in the sense that we cluster the inner train data to obtain the representation of inner test instances based on that clustered representation.

\noindent \textbf{Step 2- Learning the clustered representation of the inner train}: In each round of the inner training process, the \textit{inner train} data is fed as input to the deep embedded clustering (DEC) \citep{xie2016unsupervised}. DEC clusters the inner train instances. Then, the decoder of DEC which is trained with the inner train instances and the centroids of the clusters are used for the next steps. 

\noindent \textbf{Step 3- Extracting the latent features of inner test:} The inner test instances are given as input to the trained decoder of the previous step to obtain the latent features of inner test. 

\noindent\textbf{Step 4- Feature learning:} In this step, the clusters' centroids of the inner train which were calculated in step 2 and the latent features of inner test which were obtained in step 3 are used here to measure several features for the one inner test instance. The relations between the latent features of the inner test and the clusters' centroids of the inner train are calculated and considered as new features for that inner test sample. Thus, we extract a new set of features for each inner test instance during each run of inner folding process.



In continue, we describe each of these steps in detail.

\subsection{Step 1- Inner folding}

Inner folding provides a framework to obtain the evaluation of the representation of instances based on each other. Inner folding partition the data into $r$ non-overlapping folds, in which one fold as the \textit{inner test} is evaluated against the remaining $r-1$ folds called \textit{inner train} instances.  The inner folding process is similar to the $k$-fold cross-validation in the sense that during each round, one data fold is considered as the test sample and the remaining folds are the training samples until all the samples are considered as an inner test sample one time. The goal of defining the inner folding process is different from cross-validation as it intends to learn the characteristics of the inner test samples against the inner training samples. This process is depicted in Figure. \ref{fig:Inner_folding_process}, where the one fold inner test instance of each run is colored with green. In other words, the inner folding partitions the input data to $r$ folds and provides $r$ different versions of the input data depending on which fold is the inner test.

In a formal definition, the inner folding is an indexing function that allocates each instance, $x_i \in X$ to a $fold_{j} \in Fold$ in which $Fold=\langle fold_1,\cdots,fold_r\rangle$ by randomization. $f_{inner\ folding}: X \mapsto Fold $, in which $\forall i,j\ that\ i\neq j \rightarrow fold_{i} \cap fold_{j}= \emptyset$.
\[ f_{inner\ folding} : \{x_1, \cdots,x_n\} \mapsto \{fold_1, \cdots,fold_r\}\]
The inner folding is performed in $r$ runs as a loop where in the $J^{\text{th}}$ loop of $r$ run, the fold $J$ is considered as the inner test and the remaining $r-1$ folds make up the inner train. Similar to the cross-validation procedure, it is assumed that $\forall J,\ P_{X} \sim P_{X^{J}_{inner\ train}}$, in which $P$ refers to the distribution of data.\\

We should note that the inner folding process for the clustering task will be applied to the entire data set (i.e., $X$), however, during the IFL for the classification task, this inner folding process is only applied on the training data $X^{Train}$ (see Algorithm.\ref{alg:IFL_class_train}). Inner folding provides features for all train instances through runs on different inner test folds. For test data, we can just evaluate test instance on the whole of training data.

\begin{algorithm}
\caption{Inverse feature learning for clustering.}\label{alg:IFL_cluster}
\begin{algorithmic}[1]

\BlankLine

\STATE \textbf{Input}$:$ $X$.
\STATE \textbf{Output}$: $ \textit{Features of error representation for $X$}.

\STATE{ \textbf{Step 1: Inner folding}} %
// Partition $X$ to $r$ folds and perform $r$ runs, where in each round, $r-1$ folds are set as the inner training samples and the remaining one fold is set as inner test instances.   

\FOR{$j=1:r$}
        \STATE{Inner test = $J^{th}$ fold of $X$ --- $X^{J}_{Inner\ test}$.}
        \STATE{Inner train = All folds except $J^{th}$ fold of $X$ --- $X^{J}_{Inner\ train}$.}
        \STATE{\textbf{Step 2: Learning the clustered representation of the inner train}} 
         \STATE{${}$\hspace{0.1cm} \textbf{Input of step 2}$:$ Inner train, $X^{J}_{Inner\ train}$.}
         \STATE {${}$\hspace{0.1cm} \textbf{2.1)} Applying DEC on inner train, $X^{J}_{Inner\ train}$.}
         \STATE {${}$\hspace{0.1cm} \textbf{Outputs of step 2}$:$ Encoder of DEC, $f_{\phi}$, clusters,$\{c_i\}_{i=1}^{s}$ , and centroid of clusters --- $\{\mu_i\}_{i=1}^{s}$.}\\
         \STATE{\textbf{Step 3: Extracting the latent features of inner test}} 
         \STATE{${}$\hspace{0.1cm} \textbf{Inputs of step 3}$:$ Encoder of DEC, $f_{\phi}$, obtained in step 2, and inner test --- $X^{J}_{Inner\ test}$.}
         \STATE {${}$\hspace{0.1cm} \textbf{3.1)} Feeding encoder with inner test, $X^{J}_{Inner\ test}$.}
         \STATE {${}$\hspace{0.1cm} \textbf{Output of step 3}$:$ Latent features of inner test --- $Z^{J}_{Inner\ test}$.}
         
         \STATE {\textbf{Step 4: Feature learning}}  
         \STATE{${}$\hspace{0.1cm} \textbf{Inputs of step 4}$: $  Clusters, $\{c_i\}_{i=1}^{s}$, and their centroid, $\{\mu_i\}_{i=1}^{s}$, and latent features of inner test, $Z^{J}_{Inner\ test}$}.
          \STATE {${}$\hspace{0.1cm} \textbf{4.1)} Calculating confidence, and weight for  the inner test instances.}
         \STATE{${}$\hspace{0.1cm} \textbf{Output of step 4}$: $ Features of error representation for the inner test or $J^{th}$ fold of $X$ --- $X^{J}_{Inner\ test}$}.
 \ENDFOR     
\end{algorithmic}
\end{algorithm}

\begin{algorithm}
\caption{Inverse feature learning for classification.}\label{alg:IFL_class}
\begin{algorithmic}[1]
\STATE \textbf{Input}$:$ Train ($X^{Train}$, $Y^{Train}$), Test ($X^{Test}$).
\STATE \textbf{Output}$: $ \textit{Features of error representation for $X^{Train}$ and $X^{Test}$ }.\BlankLine

\STATE{Inverse Feature learning for the training data in classification --- See Algorithm.\ref{alg:IFL_class_train}} 
\STATE{Inverse Feature learning for the test data in classification --- See Algorithm.\ref{alg:IFL_class_test}} 

\end{algorithmic}
\end{algorithm}

\begin{algorithm}
\caption{Inverse feature learning for training data in classification.}\label{alg:IFL_class_train}
\begin{algorithmic}[1]

\BlankLine
\STATE{\textbf{1. Inverse feature learning on training data:}}
    \STATE \textbf{Input}$:$ Training data ($X^{Train}$, $Y^{Train}$).
    \STATE \textbf{Output}$: $ \textit{Features of error representation for $X^{Train}$}.

\STATE{ \textbf{Step 1: Inner folding}} %
// Partition $X^{Train}$ to $r$ folds and perform $r$ runs, where in each round, $r-1$ folds are set as the inner training samples and the remaining one fold is set as inner test instances.

\FOR{$j=1:r$}
        \STATE{Inner test= $J^{th}$ fold of $X^{Train}$ --- ${X^{Train}}^{\ J}_{Inner\ test}$.}
        \STATE{Inner train=  All folds except $J^{th}$ fold of $X^{Train}$ --- ${X^{Train}}^{\ J}_{Inner\ train}$.}
        \STATE{\textbf{Step 2: Learning the clustered representation of the inner train}} 
         \STATE{${}$\hspace{0.1cm} \textbf{Input of step 2}$:$ Inner train, ${X^{Train}}^{\ J}_{Inner\ train}$.}
         \STATE {${}$\hspace{0.1cm} \textbf{2.1)} Applying DEC on inner train, $X^{Train}{^{\ J}_{Inner\ train}}$.}
         \STATE {${}$\hspace{0.1cm} \textbf{Outputs of step 2}$:$ Encoder of DEC, $f_{\phi}$, clusters,$\{c_i\}_{i=1}^{s}$ , and centroid of clusters --- $\{\mu_i\}_{i=1}^{s}$.}
         \STATE{\textbf{Step 3: Extracting the latent features of inner test }} 
         \STATE{${}$\hspace{0.1cm} \textbf{Inputs of step 3}$:$ Encoder of DEC, $f_{\phi}$, obtained in step 2, and inner test --- ${X^{Train}}^{\ J}_{Inner\ test}$.}
         \STATE {${}$\hspace{0.1cm} \textbf{3.1)} Feeding encoder with inner test, ${X^{Train}}^{\ J}_{Inner\ test}$.}
         \STATE {${}$\hspace{0.1cm} \textbf{Output of step 3}$:$ Latent features of inner test--- $Z^{J}_{Inner\ test}$.}
         
         \STATE {\textbf{Step 4: Feature learning}}  
         \STATE{${}$\hspace{0.1cm} \textbf{Inputs of step 4}$: $ Clusters, $\{c_i\}_{i=1}^{s}$, and their centroid, $\{\mu_i\}_{i=1}^{s}$, and latent features of inner test, $Z^{J}_{Inner\ test}$}.
          \STATE {${}$\hspace{0.1cm} \textbf{4.1)} Calculating confidence, weight, and accuracy for $J^{th}$ fold.}
         \STATE{${}$\hspace{0.1cm} \textbf{Output}$: $ Features of error representation for the inner test or $J^{th}$ fold of $X^{Train}$ --- ${X^{Train}}^{\ J}_{Inner\ test}$ }.
 \ENDFOR     
\end{algorithmic}
\end{algorithm}

\begin{algorithm}
\caption{Inverse feature learning for test data in classification.}\label{alg:IFL_class_test}
\begin{algorithmic}[1]

\BlankLine
\STATE{\textbf{2. Inverse feature learning on test instances:}}
    \STATE \textbf{Input}$:$ Training data ($X^{Train}$, $Y^{Train}$) and Test data ($X^{Test}$).
    \STATE \textbf{Output}$: $ \textit{Features of error representation for $X^{Test}$}.

\STATE{\textbf{Step 1: Inner folding does not applied for test data.}}
\STATE{\textbf{Step 2: Learning the clustered representation of the training data}} 
\STATE{${}$\hspace{0.1cm} \textbf{Input of step 2}$:$ Training data, $X^{Train}$.}
 \STATE {${}$\hspace{0.1cm} \textbf{2.1)} Applying DEC on training data, $X^{Train}$.}
 \STATE {${}$\hspace{0.1cm} \textbf{Outputs of step 2}$:$ Encoder of DEC, $f_{\phi}$, clusters,$\{c_i\}_{i=1}^{s}$ , and centroid of clusters --- $\{\mu_i\}_{i=1}^{s}$.}
 \STATE{\textbf{Step 3: Extracting the latent features of test data }} 
 \STATE{${}$\hspace{0.1cm} \textbf{Inputs of step 3}$:$ Encoder of DEC, $f_{\phi}$, obtained in step 2, and test data, $X^{Test}$.}
 \STATE {${}$\hspace{0.1cm} \textbf{3.1)} Feeding encoder, $f_{\phi}$, with $X^{Test}$.}
 \STATE {${}$\hspace{0.1cm} \textbf{Output of step 3}$:$ Latent features of $X^{Test}$ --- $Z_{Test}$.}
 
 \STATE {\textbf{Step 4: Feature learning}}  
 \STATE{${}$\hspace{0.3cm} \textbf{Inputs of step 4}$: $ Clusters, $\{c_i\}_{i=1}^{s}$, their centroid, $\{\mu_i\}_{i=1}^{s}$ of $X^{Train}$, and latent features of test data, $Z_{test}$}.
  \STATE {${}$\hspace{0.3cm} \textbf{4.1)} Calculating confidence, weight, and accuracy for test data --- $X^{Test}$.}
 \STATE{${}$\hspace{0.3cm} \textbf{Output}$: $ Features of error representation for test data --- $X^{Test}$}.
   
\end{algorithmic}
\end{algorithm}

\subsection{Step 2- Learning the clustered representation of the inner train} \label{subsec:step2_details}
To handle the data sets with a large number of input features, the deep embedded clustering \citep{xie2016unsupervised} is selected in this paper to perform the clustering of inner instances, since this method is based on autoencoder that transforms the raw data to a latent feature space with a much smaller number of features. As shown in Figure \ref{fig:DEC}, the DEC is based on two phases. In the first phase, the autoencoder is trained based on the reconstruction loss on input data. In the second phase, the decoder and the centroid of latent features are given as input to Kullback–Leibler (KL) divergence minimization to improve the clustering performance. In the following, we first describe how the DEC works and then how we use the DEC in step 2 and its outcomes in step 3 and step 4 of the IFL.

The autoencoder is an unsupervised data representation that minimizes the reconstruction loss through the network on the input data. The autoencoder network consists of an encoder and a decoder. The encoder is considered as a function $z=f_{\phi}(x)$ that maps the input $x$ into a latent representation $z$ and the decoder is a function that reconstructs the latent representation to input $x'=g_{\theta}(z)$.  Parameters $\phi$ and $\theta$ denote the weights of the encoder and decoder networks, receptively and are learned through the minimizing reconstruction loss --- Figure.\ref{fig:diagram_DEC_ae}. We used the mean square error as the distance measure  \citep{xie2016unsupervised}; thus, the optimization objective is defined as follow:
\[\displaystyle\min_{\phi, \theta}L_{rec}  = min\frac{1}{n} \displaystyle\sum_{i=1}^{n}|| x_i -g_{\theta}(f_{\phi}(x_{i}))||^{2}.\]

The second phase of DEC is clustering with KL divergence which is based on two procedures that are repeated alternately till the clustering method converges. The first procedure, called procedure 2.1 here, computes a soft assignment between the embedded points and the centroids of the clusters. The second procedure, procedure 2.2, updates the weights of the encoder, $f_{\phi}$, and uses auxiliary target distributions to learn from the current assignment to refine the cluster centers. 

\noindent\textbf{Procedure 2.1: Soft Assignment}

The probability of assignment, $q_{ij}$, as a soft assignment for a sample $i$ to a cluster $j$ is calculated by Student's t-distribution \citep{maaten2008visualizing} as a kernel that measures the similarity between embedded point $z_i$ and centroid $u_j$:

\[  q_{ij} = \frac{(1+||z_{i}-\mu_{j}||^{2}/\alpha)^{-\frac{\alpha+1}{2}}}{\sum_{j'}(1+||z_{i}-\mu_{j'}||^{2}/\alpha)^{-\frac{\alpha+1}{2}}}
\]
where $z_i = f_{\phi}(x_i) \in Z$ denote the latent feature of $x_i$, $\alpha$ is the degree of freedom of Student's t-distribution. Following \citep{xie2016unsupervised}, we consider $\alpha=1$ in all experiments.

\noindent\textbf{Procedure 2.2: KL Divergence Minimization}

In this step, KL divergence loss is defined to train the model by matching the soft assignment, $q_i$ of the procedure 2.1, to the auxiliary target distribution, $p_j$, as follow:
\[L= KL(P||Q)= \displaystyle\sum_{i}\sum_{j}p_{ij}log\frac{p_{ij}}{q_{ij}}\]

We follow the defined target distribution of \citep{xie2016unsupervised}:

\[ p_{ij} = \frac{q_{ij}^2/f_j}{\sum_{j'} q_{ij'}^{2}/f_{j'}}\]
where $f_j = \sum_{i}q_{ij}$ are soft cluster frequencies. The learning is based on optimizing the cluster centers $\{\mu_{j}\}$ and the weight of network, $\theta$. 

As mentioned in the previous step, the inner folding is applied on $X^{Train}$ in classification and on $X$ in clustering resulting in $r-1$ folds as inner train and one fold as inner test. Inner train folds in each run, $J^{\text{th}}$, are given as input, $X_{inner train}^{J}$, to the DEC in which the autoencoder in the first phase is trained with inner train instances --- Figure \ref{fig:diagram_23}. Then, the latent features of inner train are learned, $Z^{J}_{inner\ train}$, and the centroid of the latent features are used as initial clusters' centroid for the clustering in the second level. The DEC uses the encoder, the initial clusters centroid, and the latent features to improve the clustering accuracy by using KL divergence which provides $\{c\}_{i=1}^{s}$ and the corresponding centroid $\{\mu\}_{i=1}^{s}$ --- in Figure.\ref{fig:diagram_23}. The encoder in the first phase of DEC is used in step 3 and the clusters and their centroid are utilized in step 4 of the IFL.

\subsection{\textbf{Step 3- Extracting the latent features of inner test}}

Now in step 3, the encoder part of autoencoder which was trained in the first phase of DEC, $f_{\phi}$, is fed with the inner test instances of the run, $X_{inner\ test}^{J}$, to obtain the corresponding latent features of inner test instances, $Z^{J}_{inner\_test}$, as shown in step 3 in Figure \ref{fig:diagram_23}.

The output of this step is a set of the latent features learned for the inner test instances of the training data, $X^{Train}$, in classification and from $X$ in clustering. It should be noted that for the test data in classification, there is not any inner train or inner test since labels are used in classification and the labels of test data in classification are not available.  The latent representation of test data in classification, $X^{Test}$, is obtained based on the autoencoder that was trained over the test data in step 2 --- see Algorithm \ref{alg:IFL_class_test}.

\subsection{\textbf{Step 4- Feature Learning}} \label{subsec:step4_details}
We consider two sets of non-overlapping instances as inner train and inner test for training data in classification or the whole instances in clustering. Also, training and test data in classification are two separate sets. The representation of one set, inner test instances or test instances, is evaluated based on the clustered representation of other set instances, inner train or training.  We assign each inner-test instance to all possible clusters and measure the representation of such assignment on the clustered representation  and map in the form of features as the corresponding error representation of that inner test instance for that cluster.
Here, we introduce three metrics which are measured per instance of each cluster, $\{c_i\}_{i=1}^s$. Two metrics of \textit{confidence} and \textit{weight} are calculated easily without the labels and can be used in both classification and clustering. Also, we extract \textit{the accuracy of assignment} for clusters by using the labels of training for classification purposes.

\textit{\textbf{Confidence}} is the ratio of the number of elements in a  cluster in which the inner-test instance has the closest distance to its center  to the  number of all instances in all clusters. Thus, we need to find the closest cluster by measuring the distance between the latent space of the instance to the centroids of the clusters which are obtained from the DEC and calculating the number of instances which belongs to that cluster to the  number of all instances. If $dist(z_j, c_b)= min (dist(z_j, \{c_i\}_{i=1}^{s}))$,  $confidence(x_{j})=\frac{|c_{b}|}{|C|}$ in which $z_j$ is the corresponding latent feature of $x_j$. Confidence is one feature. It can be used in both clustering and classification purposes.

In classification, we replicate the instances as the number of clusters and we place the learned features of each cluster in one of the replications.

\textit{\textbf{Weight}} of belonging an instance in inner test to a cluster on inner train is the distance of the latent feature of the element to the center of the cluster, $\mu$. We calculate the \textit{weight} for each inner test and the centers of all clusters, $\{\mu\}_{i=1}^{s}$, through different runs of inner folding. In other words, \textit{weight} is a vector in length of number of clusters in which $weight(x_{j}, \{\mu\}_{i=1}^{s})$= dist $( z_{j},\{\mu\}_{i=1}^{s})$. Thus, the number of features of weight is the same as the number of clusters. This measure can be used in both clustering and classification purposes.

\textit{\textbf{Accuracy of Assignment}} is the unsupervised clustering accuracy (ACC) of a cluster. It requires the labels of the instances; thus, it can be only applied in classification purposes. Thus, we use the selected cluster for each inner test, $x_{j}$, in confidence that has the closest distance $c_{d}$ and measure ACC for that cluster ---  $Accuracy\ of\ Assignment(x_{j})$ = $ACC(c_{d})$. The number of ACC feature is one.

We track the clusters through different runs of inner-folding to place the calculated \textit{weight} feature of inner test instances of clusters which are the most representative of the same elements in the same columns through different runs. It can be calculated and verified that the clusters can be tracked correctly if the ACC of a clustering approach which is used in step 2 on instances is more than $r\% + \frac{1}{s}\%$. In this paper, we set $r= 10$; thus, the ACC of the clustering approach should be at least $35\%$ for a clustering problem with 4 clusters. The ACC of the clustering approach in step 2 should be at least $20\%$ for a problem with 10 clusters. $20\%$ and $35\%$ are low standards for clustering approaches especially deep ones. DEC's ACCs on such data sets are times bigger than such thresholds. Therefore, the clusters which are representative of similar elements are tracked accurately. This process is similar to using  the labels to track the clusters as the representative of classes.

\section{Experimental Results}

In this section, we evaluate the performance of the proposed feature learning model. Since this is the first work to propose the strategy of  feature learning based on the representation of error, we consider a basic framework that can be applied across various data sets in order to demonstrate the capability of this strategy in achieving high accuracy in both classification and clustering applications when only using a small set of learned features. Indeed, this strategy can be further extended to develop refined implementations to better match the characteristics of different types of input data sets. To present a fundamental implementation of the IFL, we deployed DEC that uses an autoencoder based on fully connected networks (FCNs) \citep{xie2016unsupervised}.
Thus, DEC can be applied in diverse types of input data such as real-value, text, and some types of images.  Although the DEC or other deep clustering based on FCNs \citep{aljalbout2018clustering,min2018survey} cannot offer a comparable performance related to the models based on ConvNets for image data sets while they can be applied in a wide set of data sets. For example, deep clustering models based on ConvNets can learn the locality and shift-invariance in images much better than other architecture such as FCNs, however, the structures of ConvNets usually have limited range of usages in other domains\citep{aljalbout2018clustering, min2018survey}. We should note that the majority of the state-of-the-art deep learning methods are based on using  augmentations or generative approaches and regularization inside the process that considerably improve their performance.
As mentioned earlier, in this paper we mainly focus on the representation of error rather than attempting to develop the most optimal network architecture. Developing more advanced clustering techniques can be considered as future work. In this paper, following DEC \citep{xie2016unsupervised}, we used the autoencoder network that is based on FCNs in 9 layers that dimensions are d–500–500–2000–10-2000-500-500-d, where d is the number of features of data sets and the batch size is 256 for all of the experiments. 

\begin{table}[t]
\caption{ Descriptions of the data sets.}
\label{datasetsdesriptoin}
\begin{center}
\begin{small}
\begin{sc}
\resizebox{1\columnwidth}{!}{
\begin{tabular}{lrcr}
\toprule
\textbf{Data sets} & \textbf{\#instances} & \textbf{\#features} & \textbf{\#classes}\\ 
\midrule
\hline
MNIST    & 70000 & 26*26 (784) & 10 \\
\hline
USPS    & 9298 &  16*16 (256)& 10 \\
\hline
Reuters    & 685071 &  2000& 4 \\
\hline 
Reuters-10K   & 10000 &  2000 & 4 \\
\hline
HHAR    & 10299 &  561& 6 \\
\bottomrule
\end{tabular}}
\end{sc}
\end{small}
\end{center}
\vspace{-12pt}
\end{table}

\begin{table*}[t]
\caption{Comparison of clustering accuracy for different data sets.}
\label{table:datasets_result_clustering}
\begin{threeparttable}
\begin{center}
\begin{Large}
\resizebox{1.5\columnwidth}{!}{
\begin{tabular}{|c|c|c|c|c|c|}
\toprule
\  & MNIST & USPS & Reuters-10K  & Reuters & HHAR\\
\midrule
\hline
k-means  & 53.24 & 67.14 & 54.95  &  53.29 & 60.35\\ \hline 
HCA Average  & 19.17 & 27.95  & 44.72 & N/A &  36.12  \\\hline 
HCA Ward   & 62.11 & 74.75 & 52.62  &  N/A  &   62.02\\\hline 
LDMGI \citep{yang2010image} & 84.2 \tnote{1} & 58.0\tnote{1}   & 65.62 \tnote{3} & N/A &63.43 \tnote{3}\\ \hline 
autoencoder+k-means & 81.84 \tnote{2}  &  69.31 \tnote{4}   & 66.59 \tnote{2}  & 71.97 \tnote{2} & N/R\\ \hline 
autoencoder+SEC \tnote{2}\; \citep{nie2011spectral}  & 81.56 &  N/R  & 61.86 & N/A  & N/R \\\hline 
autoencoder+ LDMGI \; \tnote{2} & 83.98  & N/R  & 42.92 & N/A &  N/R  \\\hline 
GMMs \tnote{3} & 53.73 & N/R & 54.72 & 55.81  & 60.34 \\\hline 
autoencoder+GMM \tnote{3} & 82.18 & N/R  & 70.13 & 70.98  & 77.67 \\\hline
VAE+GMM \tnote{3}  & 72.94 & N/R  & 69.56 & 60.89 & 68.02 \\\hline
AAE\tnote{3}  \; \citep{makhzani2015adversarial}  & 83.48 &  N/R & 69.82  & 75.12  & 83.77 \tnote{3}\\\hline 
DEC \citep{xie2016unsupervised} & 84.30 & 74.08 \tnote{4}  & 72.17 & 75.63  & 79.86 \tnote{3} \\\hline 
IDEC \citep{guo2017improved} & 88.06 & 76.05  & 75.64 &  N/R  &  N/R \\\hline
VaDE \citep{jiang2017variational} & 94.46 & N/R  & 79.83 & 79.38 & 84.46 \\\hline 
IMAST(RPT) & 89.6 (5.4) & N/R & 71.9 (6.5) & N/R& N/R\\\hline 
IMAST(VAT) & \textbf{98.4} (0.4) & N/R & 71.0 (4.9) & N/R & N/R \\\hline \hline 
k-means (IFL) & 16.40 & 14.17  & 32.97(0.3) & 75.00 & 23.26 \\\hline
HCA Average (IFL )    & 82.06 & 84.80  &  76.05 & N/A&  36.50 \\\hline 
HCA Ward (IFL )    & 91.36 & 77.23 &  76.77  & N/A  &  51.60\\\hline 
DEC  (IFL)       & 91.75 & 82.68& 77.15  &  80.11 &   51.59\\\hline 
\hline 
k-means (Primary + IFL) & 91.49 & 77.44  & 77.53 & 75.56 & 51.58 \\\hline
HCA Average (Primary + IFL)    & 77.85 & \textbf{85.26}  & 78.45& N/A &  36.50 \\\hline 
HCA Ward (Primary + IFL)    & 91.26 & 77.03 & 80.08 & N/A  &   51.60\\\hline 
DEC  (Primary + IFL)       & 95.79 & 84.55& \textbf{83.19} &  \textbf{81.39}  &  \textbf{87.66}\\
\bottomrule
\end{tabular}}
\end{Large}
\end{center}
\begin{tablenotes}\footnotesize
\item [1] Results reported from \citep{ghasedi2017deep}.
\item [2] Results reported from \citep{xie2016unsupervised}.
\item [3] Results reported from \citep{jiang2017variational}.
\item [4] Results reported from \citep{guo2017improved}.
\end{tablenotes}
\end{threeparttable}

\vskip -0.1in
\end{table*} 

In this paper, we evaluated the proposed method on five common data sets \citep{xie2016unsupervised, guo2017improved, jiang2017variational,aljalbout2018clustering,min2018survey} that are described in Table \ref{datasetsdesriptoin}. These data sets cover a variety of data types, sizes, and dimensions out of which, two are handwritten digit image data sets (USPS \footnote{http://www.cs.nyu.edu/˜roweis/data.html}, MNIST \citep{lecun1998gradient}), and two of them include extracted features of documents ( Reuters-10K \citep{lewis2004rcv1} and Reuters \citep{lewis2004rcv1}), and the other one is HHAR \citep{stisen2015smart} which is a multi-class benchmark with real values of the sensory activities.  The details of the data sets and pre-processing of them are described in the following:

\textbf{MNIST:} MNIST data set \citep{lecun1998gradient} include 70000 handwritten digits with size 28*28 pixels that range of the values is between [0, 255]. The value of each pixel divided into max values to transform into an interval [0, 1]. Since we are using FCN, we reshape that in form of a vector in size 784 \citep{xie2016unsupervised}.

\textbf{USPS:} USPS data set contains 9298 gray-scale handwritten digits with size 16*16 pixels that the range of the values is between [-1, 1]. The value of each pixel divided into two to transform into an interval [-0.5, 0.5].

\textbf{Reuters:} Reuters data set \citep{lewis2004rcv1} includes 810000 English news stories that are labeled based on a category tree. We followed \citep{xie2016unsupervised} to use 4 root categories: corporate/industrial, government/social, markets, and economics as labels while the documents with multiple labels have not considered. The result documents are 685071 ones that tf-idf features are computed on the 2000 word stems with most frequent occurring. 

In \textbf{Reuters-10K}, the 10000 random documents of Reuters are sampled as same as \citep{xie2016unsupervised}. 

\textbf{HHAR}: The Heterogeneity Human Activity Recognition (HHAR) data sets that contains sensory information about smartphones and smartwatches. The data set include 10299 instances with 561 features of 6 categories of human activities. we follow the data set as same as \citep{jiang2017variational}.

Here, we compare the performance of the proposed IFL with some state-of-the-art techniques that are fairly general such as \citep{xie2016unsupervised, guo2017improved, jiang2017variational} in clustering or several common classification methods which can handle real values, text or images. In this paper, we set the number of runs for the inner folding process to 10 (i.e., $r=10$) in both classification and clustering for all experiments. We repeat the experiments 5 times to calculate the variance of the results in both classification and clustering. The numbers in the parenthesis show the variance. If there are not any parentheses, it means that the variance is smaller than 0.01.

\subsection{Clustering}

\begin{figure}
  \begin{subfigure}[b]{0.49\columnwidth}
    \includegraphics[width=\linewidth,height=3cm]{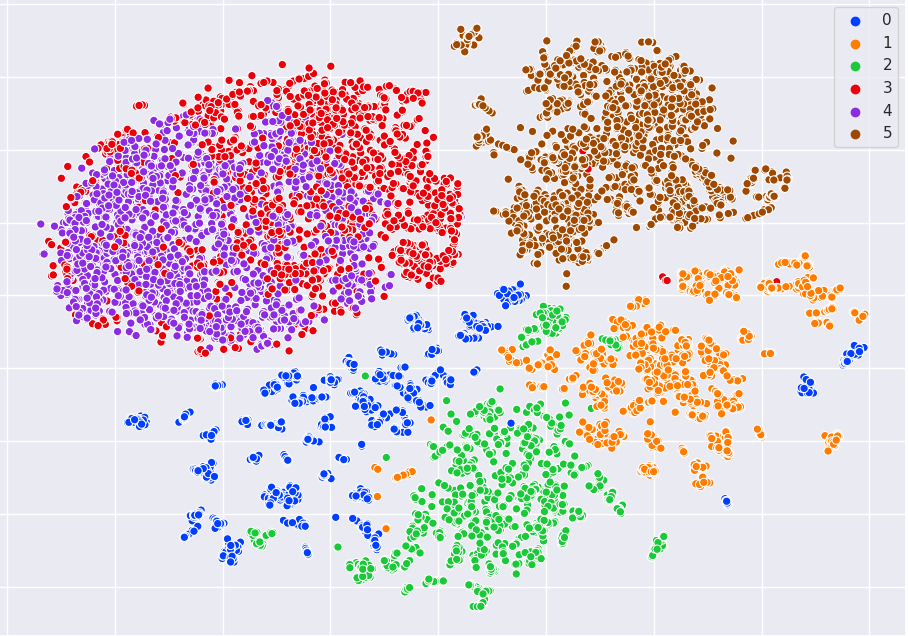}
    \caption{Representation of input.}
   \end{subfigure}
  \hfill 
  \begin{subfigure}[b]{0.49\columnwidth}
    \includegraphics[width=\linewidth,height=2.5cm]{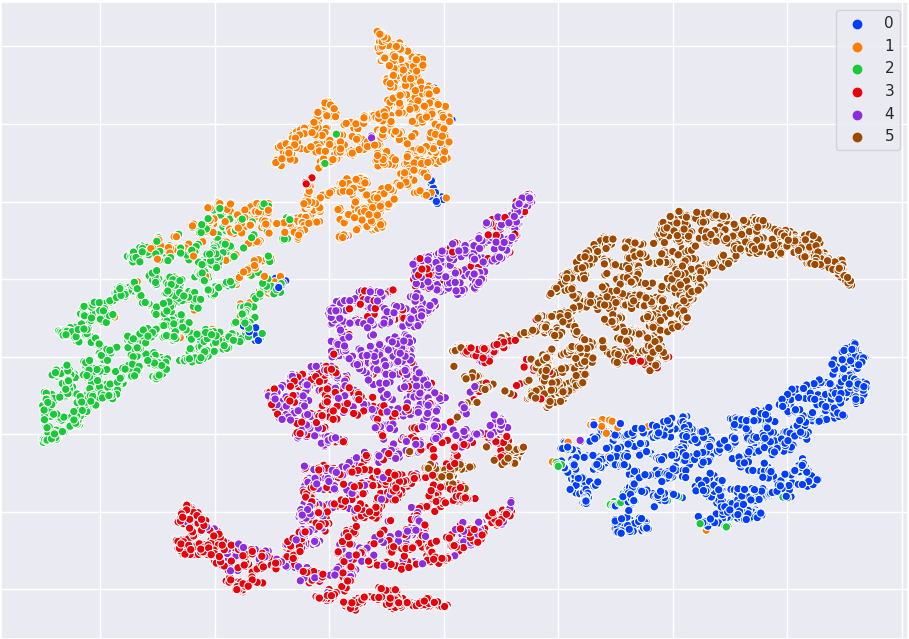}
    \caption{Iteration 0 of phase 2 of DEC, ACC $\simeq$ 70\%.}
  \end{subfigure}
  \begin{subfigure}[b]{0.49\columnwidth}
    \includegraphics[width=\linewidth,height=2.5cm]{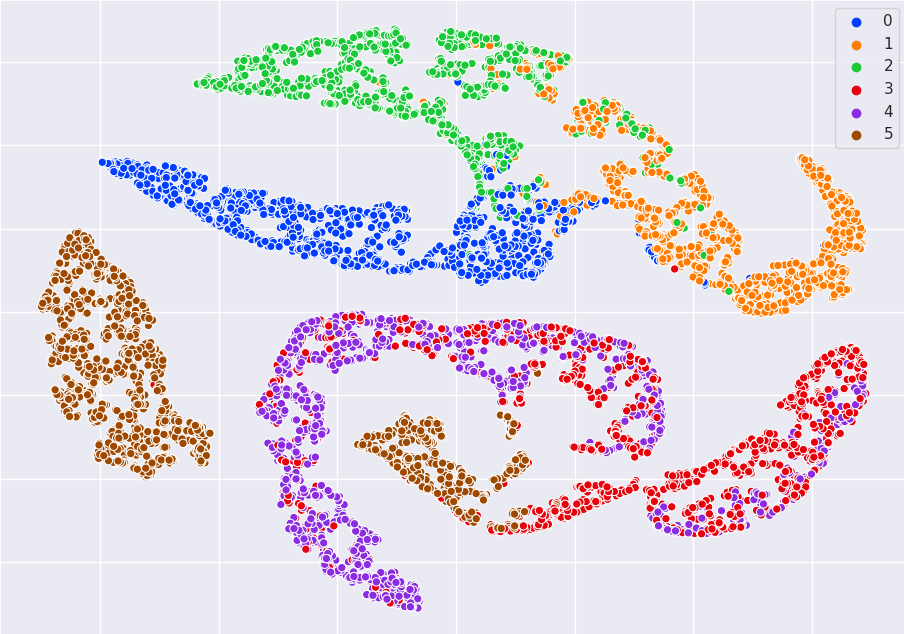}
    \caption{Iteration 360 of phase 2 of DEC, ACC $\simeq$ 80\%. }
  \end{subfigure}
  \hfill 
  \begin{subfigure}[b]{0.49\columnwidth}
    \includegraphics[width=\linewidth,height=2.5cm]{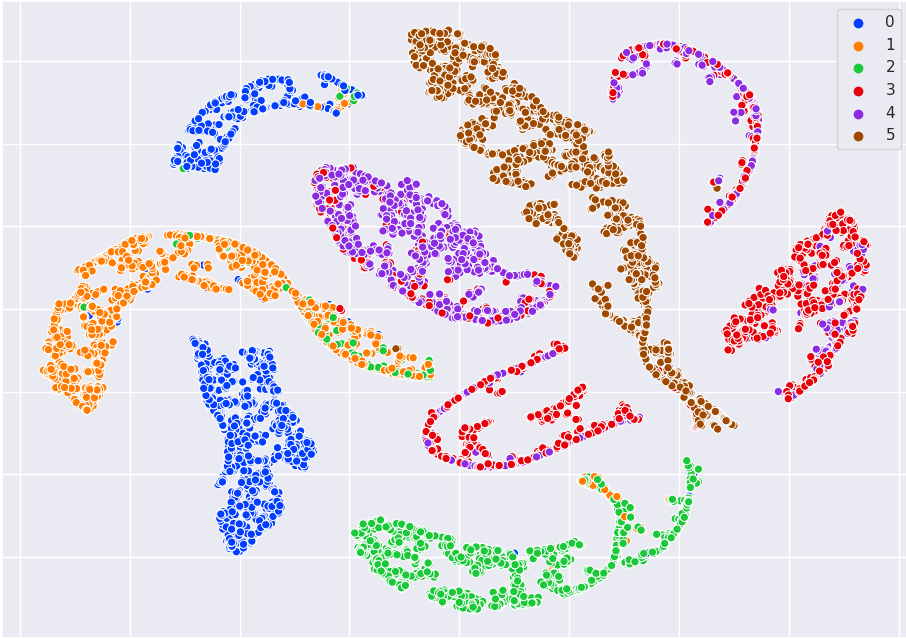}
    \caption{Iteration 1800 of phase 2 of DEC, ACC $\simeq$ 87\%.}
  \end{subfigure}
  
 \caption{ The representation of HHAR data sets while the learned features of error are added to corresponding instances by t-SNE through different phases of DEC.}
\label{fig:T_sne_diagram}
\vspace{-0.4 cm}  
  
\end{figure}

Two main categories in clustering methods are the classical clustering and deep clustering. Classical clustering methods include partition-based method (e.g. K-means), spectral clustering such as density-based clustering methods, and  hierarchical clustering (e.g., agglomerative clustering). The spectral clustering methods are not scalable to large data sets considering their computational complexity \citep{xie2016unsupervised}. The deep learning based clustering methods are developed based on different criteria such as the architecture or the loss as reviewed in \citep{aljalbout2018clustering,min2018survey}. The clustering methods with deep learning have shown superior performance in several applications (e.g., ConvNets in image data sets, or the generative-based models). Here, we implement the idea of  inverse feature learning using DEC that shows our results are comparable or even better than the state-of-the-art methods, where we select at least one clustering method out of each category.

The results are compared with classical clustering methods such as k-means\citep{macqueen1967some}, hierarchical cluster analysis (HCA) based on bottom up approach for the ward and average linkage criterion \citep{hastie2009elements}, and Gaussian mixture model (GMMs) as well as several deep learning based clustering methods including LDMGI \citep{yang2010image}, autoencoder+k-means, autoencoder + LDMGI \citep{yang2010image}, autoencoder+SEC \citep{nie2011spectral}, DEC, improved DEC (IDEC) \citep{guo2017improved}, VaDE \citep{jiang2017variational} which is a generative and based on variational autoencoder (VAE) \citep{kingma2013auto}, autoencoder+GMM, VAE+GMM, adversarial autoencoder (AAE) \citep{makhzani2015adversarial}, and information maximizing
self-augmented training (IMAST) \citep{hu2017learning}. We do not compare the results with the clustering approaches based on ConvNets as they can just be applied on image data sets.

The common unsupervised clustering accuracy, denoted by ACC, is used to evaluate the performance of this model as defined as follows:
\begin{equation}
  \textit{ACC}=\max_{m}\frac{\sum_{i=1}^{n} 1\{y_i=m(c_i)\}}{n}
\end{equation}
where $c_i$ is the cluster assignment for instance $i$ determined by the clustering approach, $y_i$ is the \textit{ground-truth} label or the label of $i^{\text{th}}$ instance, and $m$ is the mapping function that ranges over all possible one-to-one mapping between $c_i$ and $y_i$. 
 
The ACC of the clustering models is reported for the primary features, just with the learned features of error (IFL), and the primary features in addition to the learned features (Primary +IFL). It should be noted that the IFL features are highly abstract, (i.e., only $1+s$ in which $s$ shows the number of clusters). This number of features is significantly less than the number of original features of data sets --- several hundred times. For instance, the number of IFL features in Reuters and Reuters 10k data set is $5$ while the number of primary features is $2000$. In MNIST and USPS, the numbers of primary features are $784$ and $256$, respectively while the number of IFL features is $11$. For HHAR data set, the number of primary features is $561$ while IFL features are just $7$ ones.

We evaluate IFL based on two scenarios to show how the IFL feature works separately and along with the primary features in Table \ref{table:datasets_result_clustering}. HCA clustering approaches and LDMGI cannot be applied to Reuters data set. We use the open libraries of scikit-learn \citep{scikit-learn} for HCA that cannot handle the data sets like Reuters. Also, as motioned in \citep{xie2016unsupervised} LDMGI and SEC need ``months of computation time and terabytes of memory'' for Reuters.

As shown in Table \ref{table:datasets_result_clustering}, the approaches such as AAE which is based on GAN autoencoder, variational deep embedding (VaDE) which is a generative model based on VAE or IMAST (VAT) which uses regularization and self-augmentation provide better performance compared to other techniques. It can be seen in the table that IMAST (VAT) provides the best results on MNIST data set (even better than ConvNets \citep{hu2017learning}), however. it cannot keep the superiority over other data sets. In fact, the self-training augmentation of IMAST helps the networks handle shift-invariance issues in image data sets, but in data sets such as Reuters 10K that this issue does not exist, its performance in comparison to other deep learning methods degrades considerably. VaDE which utilizes both variational autoencoders and generative models shows a robust superior performance related to other approaches except on MNIST data set. HHAR data set is not a sparse data set like other data sets that makes it a challenging one for classic clustering approaches.

Table \ref{table:datasets_result_clustering} shows when the IFL features are used in addition to the original features in different clustering methods such as DEC and k-means, they provide robust results with a negligible variance (i.e., smaller than 0.01).  In this table, ``N/R'' to the cases, for which the results were not reported and ``N/A'' refers to the cases where a particular approach cannot be applied one a data set because of the required memory or the required time as mentioned in \citep{xie2016unsupervised}.

In the first scenario, we only used the IFL features alone, and noting that the number of INF features is considerably less than the primary features, the k-means clustering method cannot work properly on most of the data sets except Reuters, in which there are considerable instances. HCA based on average and ward linkage offers interesting results in comparison to other deep learning methods except for the HHAR data set. The results of this table show that the IFL features are not descriptive enough to be used alone on the data sets which are not sparse. However, the performance of the DEC approach using only the IFL features is considerably better than the cases when this clustering is applied on the primary features or many other deep clustering methods including LDMGI, autoencoder +(k-means, SEC, LDMGI, GMM), VAE+GMM, AAE or IDEC.

In the second scenario, the learned features are used along with the primary features for DEC or other clustering approaches. In this scenario, we achieved the best results compared to several state-of-the-art techniques, except for the MNIST data set, where our best performance is about 2\% lower than IMAST --- as we mentioned earlier, IMAST uses self-training augmentation to handle shift-invariance of images and its results are not steady in other types of data sets such as Reuters 10k, where the accuracy of IMAST is 10\% lower than our proposed approach. Moreover, the results in Table \ref{table:datasets_result_clustering} demonstrate that adding the IFL features to the original features considerably improves the performance of the classical clustering techniques such as k-means and HCA to the point that they achieve comparable results with the state-of-the-art deep clustering methods.

We demonstrate the latent representations of input data and latent feature space for HHAR data set through different phases of DEC into 2D space by t-SNE \citep{maaten2008visualizing} in Figure \ref{fig:T_sne_diagram}. As it can be seen in this figure, the clusters become well separated through the steps while the accuracy improves. 

In summary, based on the extensive experimental results, we can conclude that the IFL features provide a new perspective that offers considerable accuracy improvement in DEC and even the classic clustering techniques. Even when the IFL features are fed alone to the clustering methods, they provide a proposer representation of the data set that improves the performance of the  classic clustering techniques to the level of deep-learning-based clustering approaches. Thus, the IFL framework can provide a practical and fast solution to enhance the performance of the classical clustering techniques, in particular for sparse data sets such as image and text data sets.



\begin{table*}[t]
\caption{It is compared accuracy of the classifiers without and with the inverse feature learning based on the first technique. In the proposed method, the learned features are independent of the classifier, hence using a better classifier can achieve higher accuracy over our learned features as it does over other sets of features.}
\label{table:datasets_result_classsfication_whole}
\vskip 0.15in
\begin{threeparttable}
\begin{center}
\begin{small}

\resizebox{2\columnwidth}{!}{
\begin{tabular}{|l||c|c||c|c||c|c||c|c||c|c|}
\toprule
\ Classifiers & \multicolumn{2}{c||}{Decision Tree}&  \multicolumn{2}{c||}{Random Forests}& \multicolumn{2}{c||}{XGboost}& \multicolumn{2}{c||}{KNNs}& \multicolumn{2}{c|}{MLP}\\

& Primary & Primary+IFL & Primary & Primary+IFL & Primary & Primary+IFL & Primary & Primary+IFL & Primary & Primary+IFL \\
\midrule
 MNIST  & 88.6(0.01) & 93.8  & 96.9(0.01) & 97.05 & 93.7 & 95.3 & 96.9 & 97.1 &  97.8 & \textbf{98.87}\\
 USPS    & 85.8(0.13) & 90.20(0.01)  & 94.0(0.03) & 93.57(0.07) & 92.6& 92.6  &  95.0  & 95.0 &  94.3(0.09)& \textbf{94.54}(0.01)  \\
Reuters-10K    & 84.5(0.02) & 89.9(0.02)  & 94.0(0.06) & 95.1(0.01) & 92.3& 94.0 & 94.1 & 94.5 & 95.3(0.07)& \textbf{95.86}(0.04)\\
HHAR    & 85.5(0.38) & 86.6(0.07) & 92.7(0.21) & 93.2(0.02) & 93.4& 94.12 & 89.7 & 90.0  & 94.7(0.13)& \textbf{96.34}(0.02) \\
HHAR   & 85.5(0.38) & 97.65 \tnote{1} & 92.7(0.21) & 98.88 \tnote{1} & 93.4& 98.07 \tnote{1} &  89.7 & 96.94 \tnote{1}  &  94.7(0.13)& \textbf{99.0}\tnote{1} \\
\bottomrule
\end{tabular}
}

\end{small}
\end{center}

\begin{tablenotes}\footnotesize
\item [1] Results reported based on the second technique.
\end{tablenotes}
\end{threeparttable}
\vskip -0.1in
\end{table*}

\begin{table*}[t]
\caption{The accuracy of the classifiers are shown just with the learned features. It should be noted the features in IFL are highly abstract and much few in comparing the number of features of data sets (several hundred time fewer); for example, in Reuters data set, 2000 vs 3.}
\label{table:datasets_result_classsfication_IFL}
\vskip 0.15in
\begin{threeparttable}
\begin{center}
\begin{small}
\begin{sc}
\resizebox{1.4\columnwidth}{!}{\begin{tabular}{|l|c|c|c|c|c|}
\toprule
Classifier Model & MNIST& USPS & Reuters  & Reuters-10K  & HHAR    \\
\midrule
\hline
Decision Tree   & 86.63(0.04)  &  82.05 & 82.72  & 80.22(0.01)& 64.72\tnote{1} \\
\hline
Random forests   & 86.49 & 79.50 (4.6) & 83.88  & 80.09(0.01)& 76.12(0.13)\tnote{1}\\
\hline
XGboost    & 87.14 & 82.05 &  84.08 & 80.44 & 71.29\tnote{1}\\
\hline 
KNNs   & 84.8 & 81.50  & 83.59 &  79.04& 74.2\tnote{1}\\
\hline
MLP    & 86.85  & 82.05 & 84.15 & 80.24(0.33)&  77.51(1.22)\tnote{1}\\
\bottomrule
\end{tabular}}
\end{sc}
\end{small}
\end{center}
\begin{tablenotes}\footnotesize
\item [1] Results reported based on the second technique.
\end{tablenotes}
\end{threeparttable}

\vspace{-12pt}
\end{table*}
\subsection{Classification}

We evaluated the performance of the IFL features in classification by using several known and state-of-the-art classification methods in the literature which can handle different types of data including Decision Tree \citep{breiman2017classification}, Random Forests \citep{liaw2002classification} as an general ensemble method, XGboost \citep{chen2016xgboost} as a known boosting method in machine learning, $K$-nearest neighbors (KNNs), and a Multi-layer Perceptron classifier (MLP) that utilizes \textit{adam} as weight optimization. We use the libraries of scikit-learn \citep{scikit-learn} for the mentioned classifiers. Accuracy is used as the evaluation metric.

In classification, IFL features can be used based on two techniques. In the first technique, we consider the \textit{weight} feature of each cluster for each instance separately. In other words, we form a new data set that the number of its instances is the number of instances of the original data set that is multiplied with the number of clusters and the number of IFL features equals 3 --- one feature for \textit{confidence}, one feature for \textit{weight}, and one feature for \textit{accuracy of assignment}. When the IFL features are considered along with the primary features in the first technique, there are versions as the number of clusters of each instance that the primary features and also the \textit{confidence} and \textit{accuracy of the assignment} are the same among them while the \textit{weight} feature is different. The labels of all the versions of each instance are the same for training. The final label that is considered for each instance of test data is the label with the maximum value among its versions. We used the first technique for the data sets that are sparse like images and text. The second technique is like the procedure of IFL features in clustering. In the second technique, we consider the features of each instance including \textit{weight} corresponding to different clusters as a whole. Thus, the number of IFL features equal with the number of clusters+2 --- one feature for \textit{confidence}, one feature for \textit{accuracy of assignment}, and $s$ features for the \textit{weight} in which $s$ is the number of classes. We found this technique is proper for the data sets that are not sparse like HHAR. The first technique is the reliable and robust one and we consider that as the default technique. The second technique is recommended for data sets that are not sparse. It can be seen in Table \ref{table:datasets_result_classsfication_whole} for HHAR data set the second technique provides much better results in comparison to the first technique.

It should be noted that the IFL features are highly abstract, (i.e., only 3 in the first technique and only $2+s$ in which $s$ shows the number of clusters in the second technique). This number of features is significantly less than the number of original features of data sets --- several hundred times. For instance, the number of IFL features in Reuters and Reuters 10k data set is $3$ in the first technique while the number of primary features is $2000$. In MNIST and USPS, the numbers of primary features are $784$ and $256$, respectively while the number of IFL features is $3$. For HHAR data set, the number of primary features is $561$ while IFL features are just $8$ ones in the second technique. 

As shown in Table \ref{table:datasets_result_classsfication_whole}, the IFL features improved the accuracy of the Decision Tree classifier on different data sets and also bring down the variance of the results compared to the scenario when only the primary features are used. We have not reported the results on Reuters data set for classification except the MLP classifier that results for the primary features and (primary features + IFL) are 97.41\% \& 98.12\%  correspondingly, since the required time and memory were out of the capacity of our systems and the libraries. Since the decision tree directly works with the feature space, the IFL features in different data sets improve the results from 2\% to 12\%. For the Random Forests classifier, the IFL features improve the results over all data sets  except the USPS, where the accuracy is degraded for 0.03\%. For the XGboost and KNNs classifiers, the features improve the results or at least offer a comparable accuracy. In MLP, we can see a slight improvement in the results in which the variance also decreased. In conclusion, the IFL features enhance the classifiers' accuracy, especially on approaches in which the features are used directly. As shown in Table \ref{table:datasets_result_classsfication_IFL}, the IFL features when used alone provide some interesting results. It shows the IFL features capture informative aspects of data.

\section{Conclusion}

In this paper, we introduce \textit{error representation learning} as a novel perspective about the error in machine learning and \textit{inverse feature learning} as a representation learning strategy that stands on deep clustering  to learn the representation of error in the form of high-level features. The strategy of inverse feature learning can be implemented by different learning techniques. We applied the strategy based on DEC which is a general and basic clustering approach, however, the IFL framework can be applied on other clustering techniques as well. The performance of the IFL framework is evaluated on five popular data sets using different clustering techniques. We consider two evaluation scenarios, where the IFL features are used alone or in addition to the primary features. The experimental results show that even in the cases that the small set of IFL features are used, the clustering accuracy for some clustering techniques (e.g., DEC) is comparable to several other approaches. This confirms that the IFL features are highly informative. The results of the second evaluation scenario, in which the IFL features are used along with the primary ones, demonstrate considerable improvement for DEC and other classical clustering approaches. In this scenario, we achieved the best result compared to several state-of-the-art techniques over the Reuters-10K, Reuters, and HHAR data sets. It should be noted that several of these state-of-the-art approaches are generative, based on VAE, and also use self-augmentation training and customized regularization. Moreover, the experimental results show the performance of the proposed IFL framework  on a variety of known classification methods and in most cases the result is improved. Learning error representing features based on the generative and specialized network tailored to the data sets (e.g., ConvNets for images) can be considered as future works.

\bibliographystyle{IEEEtran}
\bibliography{bibio}


\end{document}